\documentclass[letterpaper, 10 pt, conference]{ieeeconf}  

\IEEEoverridecommandlockouts                              

\usepackage{cite}
\usepackage{amsmath,amssymb,amsfonts,bm}
\usepackage{algorithmic}
\usepackage{graphicx}
\usepackage{textcomp}
\usepackage{xcolor}
\usepackage[colorlinks=true, allcolors=blue]{hyperref}
\usepackage{float}
\usepackage{booktabs}
\usepackage{multirow}
\def\BibTeX{{\rm B\kern-.05em{\sc i\kern-.025em b}\kern-.08em
    T\kern-.1667em\lower.7ex\hbox{E}\kern-.125emX}}

\title{\LARGE \bf AFPN: Asymptotic Feature Pyramid Network for Object Detection}

\author{
Guoyu Yang\textsuperscript{1}, Jie Lei\textsuperscript{1*}, Zhikuan Zhu\textsuperscript{1}, Siyu Cheng\textsuperscript{1}, Zunlei Feng\textsuperscript{2}, Ronghua Liang\textsuperscript{1} \\
\textit{\textsuperscript{1}College of Computer Science, Zhejiang University of Technology, Hangzhou, P.R. China} \\
\textit{\textsuperscript{2}College of of Computer Science, Zhejiang University, Hangzhou, P.R. China} \\
\{gyyang, \textsuperscript{*}jasonlei, zzkuan, sycheng, rhliang\}@zjut.edu.cn \\
zunleifeng@zju.edu.cn
}

\begin{document}

\maketitle
\thispagestyle{empty}
\pagestyle{empty}

\begin{abstract}

Multi-scale features are of great importance in encoding objects with scale variance in object detection tasks. A common strategy for multi-scale feature extraction is adopting the classic top-down and bottom-up feature pyramid networks. However, these approaches suffer from the loss or degradation of feature information, impairing the fusion effect of non-adjacent levels. This paper proposes an Asymptotic Feature Pyramid Network (AFPN) to support direct interaction at non-adjacent levels. AFPN is initiated by fusing two adjacent low-level features and asymptotically incorporates higher-level features into the fusion process. In this way, the larger semantic gap between non-adjacent levels can be avoided. Given the potential for multi-object information conflicts to arise during feature fusion at each spatial location, adaptive spatial fusion operation is further utilized to mitigate these inconsistencies. We incorporate the proposed AFPN into both two-stage and one-stage object detection frameworks and evaluate with the MS-COCO 2017 validation and test datasets. Experimental evaluation shows that our method achieves more competitive results than other state-of-the-art feature pyramid networks. The code is available at \href{https://github.com/gyyang23/AFPN}{https://github.com/gyyang23/AFPN}.

\end{abstract}


\section{Introduction}

Object detection is a fundamental problem in computer vision, aiming to detect and localize objects in images or videos. With the advent of deep learning, object detection has seen a paradigm shift, and deep learning-based methods have become the dominant approach. Ongoing research has led to the development of many new methods, indicating the potential for further advancements in this field.

\begin{figure}[htbp]
\centering
\includegraphics[width=\linewidth]{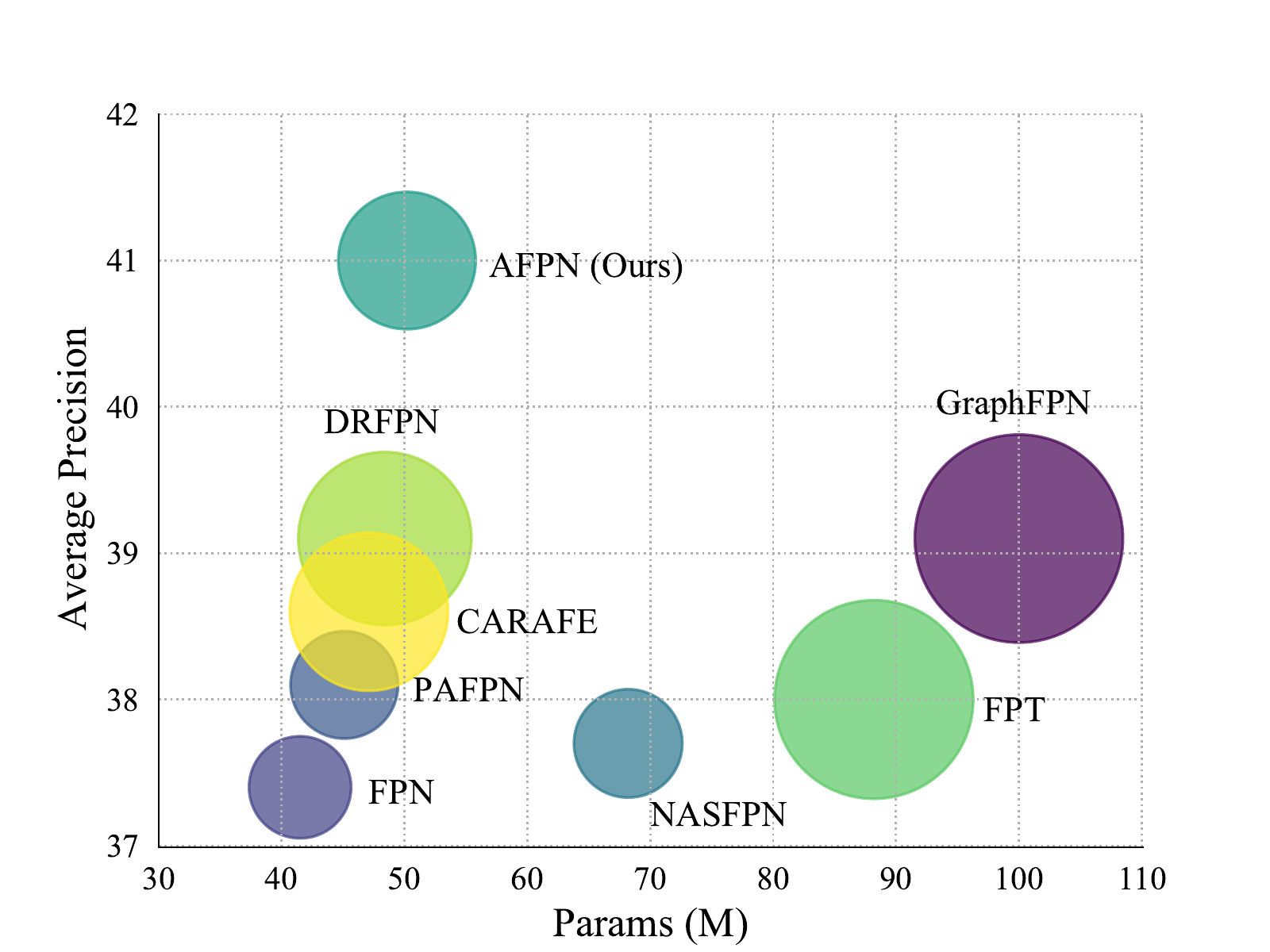}
\caption{The detection results of various feature pyramid networks on MS-COCO val2017. The area of the bubble is proportional to the GFLOPs of the method. The proposed AFPN achieves the highest AP while maintaining moderate numbers of parameters and GFLOPs.}
\label{fig:0}
\end{figure}

Object detection methods based on deep learning are typically categorized into one-stage and two-stage methods. One-stage methods~\cite{lin2017focal, yolov5, wang2022yolov7} predict the category and location of objects directly from the input image. Two-stage methods~\cite{ren2015faster, he2017mask, wu2020rethinking, zhang2020dynamic}, on the other hand, generate a set of candidate regions firstly and then perform classification and position regression on these regions. The uncertainty of the size of objects in images can lead to the loss of detailed information in feature extraction with a single scale. Therefore, object detection models usually introduce feature pyramid architectures~\cite{lin2017feature, liu2018path, ghiasi2019fpn, guo2020augfpn, zhang2020feature, zhao2021graphfpn, quan2022centralized, yang2022double} to solve the problem of scale variation. Among them, FPN~\cite{lin2017feature} is the most commonly used feature pyramid architecture. By utilizing FPN, both one-stage and two-stage detectors can achieve improved results. Based on FPN, PAFPN~\cite{liu2018path} adds a bottom-up path to the feature pyramid network, compensating for the deficiency of low-level feature details in the high-level features of FPN.

For object detection tasks, the truly useful features must contain both detailed and semantic information about the object, and these features should be extracted by a sufficiently deep neural network. In the existing feature pyramid architectures~\cite{lin2017feature, liu2018path, kirillov2019panoptic, qiao2021detectors}, high-level features at the top of the pyramid need to propagate through multiple intermediate scales and interact with features at these scales before being fused with the low-level features at the bottom. In this process of propagation and interaction, the semantic information from high-level features may be lost or degraded. Meanwhile, the bottom-up pathway of PAFPN~\cite{liu2018path} brings about the opposite problem: the detailed information from low-level features may be lost or degraded during propagation and interaction. In recent studies, GraphFPN~\cite{zhao2021graphfpn} has addressed the limitation of direct interaction between only adjacent scale features and introduced the graph neural network for this issue. However, the additional graph neural network structure significantly increases the parameters and computations of the detection model, which outweighs its benefits.

Existing feature pyramid networks typically upsample high-level features generated by the backbone network to low-level features. However, we have noticed that HRNet~\cite{sun2019deep} maintains low-level features throughout the feature extraction process and repeatedly fuses low-level and high-level features to generate richer low-level features. This approach has demonstrated outstanding advantages in the field of human body pose estimation. Inspired by the HRNet network architecture, we propose an Asymptotic Feature Pyramid Network (AFPN) to tackle the above limitations. During bottom-up feature extraction in the backbone, we initiate the fusion process by combining two low-level features with varying resolutions in the first stage. As we progress to later stages, we gradually incorporate high-level features into the fusion process, ultimately culminating in the fusion of the top features of the backbone. This fusion way can avoid the large semantic gap between non-adjacent levels. 
During this process, the low-level features are fused with the semantic information from high-level features, and the high-level features are fused with the detailed information from low-level features. Due to their direct interaction, the information loss or degradation in multi-stage transmission is avoided. 
Throughout the feature fusion process, element-wise sum is not an effective method due to there may be a contradiction of different objects in a certain position between the levels. To address this issue, we utilize adaptive spatial fusion operation to filter the features in the multi-level fusion process. This allows us to retain useful information for fusion.

To evaluate the performance of our method, we employed the Faster R-CNN framework on the MS COCO 2017 dataset. Specifically, we utilize ResNet-50 and ResNet-101 as backbones, which lead to 1.6\% and 2.6\% improvements, respectively, compared to FPN-based Faster R-CNN. We compare it against other feature pyramid networks. The experimental results indicate the proposed AFPN not only achieves more competitive results than other state-of-the-art feature pyramid networks, but also owns the lowest Floating Point Operations (FLOPs). 
Moreover, we extend the AFPN to the one-stage detector. We implement our proposed method on the YOLOv5 framework and obtain superior performance to the baseline with fewer parameters.

Our primary contributions are as follows: (1) We introduce an Asymptotic Feature Pyramid Network (AFPN), which facilitates direct feature fusion across non-adjacent levels, thus preventing the loss or degradation of feature information during transmission and interaction. (2) To suppress the contradiction of information between different levels of features, we incorporate an adaptive spatial fusion operation into multi-level feature fusion process. (3) Extensive experiments on the MS COCO 2017 validation and test datasets indicate that our method exhibits superior computational efficiency compared to other feature pyramid networks while achieving more competitive results.

\section{Related Work}

Traditional computer vision methods usually extract only one scale feature from the image for analysis and processing. This will lead to poor detection performance for objects of different sizes or scenes of different scales. Researchers have constructed feature pyramids incorporating features at various scales, overcoming the limitations of using single-scale features. Furthermore, numerous studies have proposed feature fusion modules that aim to augment or refine the feature pyramid network, further boosting the detector’s performance.

\subsection{Feature Pyramids}
FPN~\cite{lin2017feature} uses a top-down way to transfer high-level features to low-level features to achieve the fusion of different levels of features. However, high-level features do not fuse with low-level features in this process. For this reason, PAFPN~\cite{liu2018path} adds a bottom-up path based on FPN to make high-level features obtain details in low-level features. 
Unlike the fixed network architecture method, NASFPN~\cite{ghiasi2019fpn} uses the neural architecture search algorithm to automatically search for the optimal connection structure. Recently, ideas from other fields have also been introduced into the feature pyramid architecture. For example, FPT~\cite{zhang2020feature} introduces the self-attention mechanism in the NLP field to extract features at different levels and uses a multi-scale attention network to aggregate these features. 
GraphFPN~\cite{zhao2021graphfpn} uses the graph neural network to interact and propagate information on the feature pyramid. While GraphFPN also facilitates direct interaction between non-adjacent levels, its reliance on the graph neural network substantially increases parameter quantity and computational complexity, and FPT suffers from similar problems. In contrast, the AFPN only introduces normal convolutional components. Therefore, our AFPN is more feasible and practical in practical applications.

\begin{figure*}[htbp]
\centering
\includegraphics[width=0.9\linewidth]{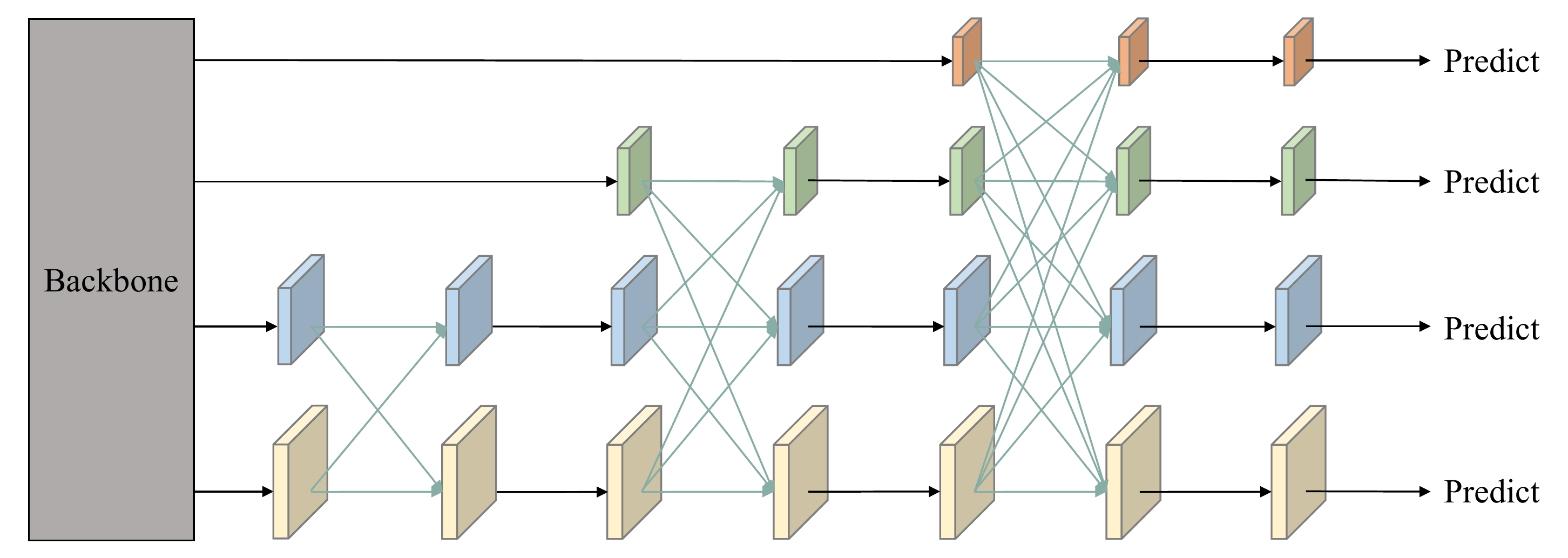}
\caption{The architecture of the proposed Asymptotic Feature Pyramid Network (AFPN). AFPN fuses two low-level features in the initial stage. The subsequent stage fuses higher-level features, while the final stage adds top-level features to the feature fusion process. Black arrows represent convolutions, and aquamarine arrows represent adaptive spatial fusions.}
\label{fig:1}
\end{figure*}

\begin{figure}[htbp]
\centering
\includegraphics[width=0.85\linewidth]{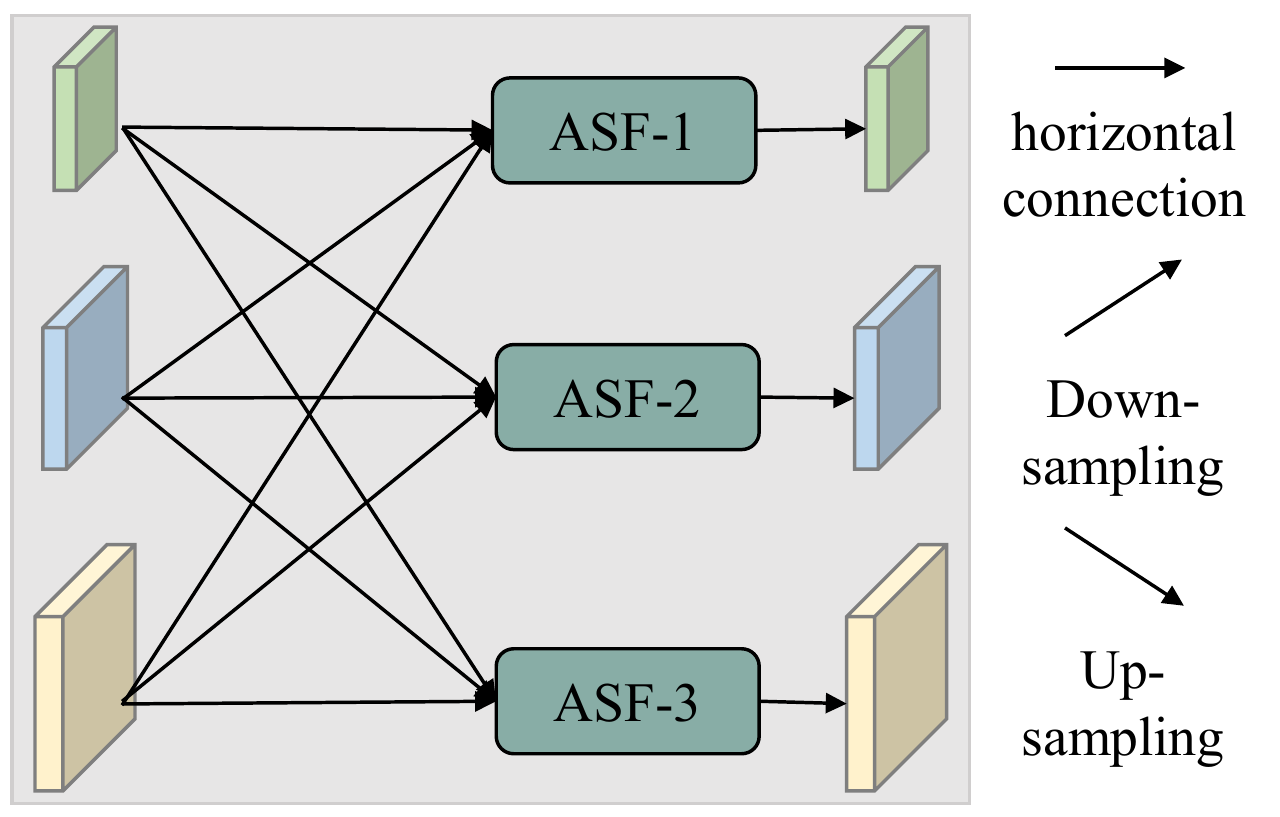}
\caption{Adaptive spatial fusion operation. This serves as an illustration of feature fusion at three different levels, but we can adapt the method as needed for cases with more or fewer levels.}
\label{fig:2}
\end{figure}

\subsection{Feature Fusion Modules}
The feature fusion module is commonly incorporated into a pre-existing, fixed-topology feature pyramid to augment its features. Several studies have also been conducted to enhance the upsampling module of the feature pyramid. In this paper, the module that does not change the topology of the feature pyramid is called the feature fusion module. CARAFE~\cite{wang2019carafe} is a universal, lightweight, and efficient upsampling operator that can aggregate large receptive field information. ASFF~\cite{liu2019learning} adds weights to features at different levels in order to fuse them effectively, given the contradictory information that may exist between features at different levels. DRFPN~\cite{ma2020dual} extends the PAFPN~\cite{liu2018path} architecture by incorporating the Spatial Refinement Block (SRB) and Channel Refinement Block (CRB). The SRB module leverages contextual information across adjacent levels to learn the location and content of upsampling points, while the CRB module utilizes an attention mechanism to learn an adaptive channel merging strategy. 
Compared to these feature pyramid architecture, the feature pyramid module can be seamlessly integrated into a wide range of existing feature pyramid architectures, providing a practical solution to address various limitations of the feature pyramid. One limitation of the feature pyramid is the co-existence of information from different objects in the same position during feature fusion. This limitation is particularly obvious in AFPN, as it requires more rounds of feature fusion. In further, We perform adaptive spatial fusion to effectively fuse features at different levels.

\section{Asymptotic Feature Pyramid Network}

\subsection{Extracting Multi-level Features}

Like many object detection methods based on feature pyramid networks, different levels of features are extracted from the backbone before feature fusion. We follow the design by Faster R-CNN~\cite{lin2017feature} framework which extract the last layer of features from each feature layer of the backbone, resulting in a set of features at different scales represented as \{$C_{2}$, $C_{3}$, $C_{4}$, $C_{5}$\}. To perform feature fusion, the low-level features $C_{2}$ and $C_{3}$ are first input into the feature pyramid network, followed by the addition of $C_{4}$, and finally $C_{5}$. Following the feature fusion step, a set of multi-scale features \{$P_{2}$, $P_{3}$, $P_{4}$, $P_{5}$\} is produced. For the experiments conducted on the Faster R-CNN framework, we apply a convolution with a stride of 2 to $P_{5}$, followed by another convolution with a stride of 1 to generate $P_{6}$, which ensures a unified output. The final set of multi-scale features is \{$P_{2}$, $P_{3}$, $P_{4}$, $P_{5}$, $P_{6}$\}, with corresponding feature strides of \{4, 8, 16, 32, 64\} pixels. It should be noted that YOLO only inputs \{$C_{3}$, $C_{4}$, $C_{5}$\} into the feature pyramid network, which generates an output of \{$P_{3}$, $P_{4}$, $P_{5}$\}.

\begin{table*}[htbp]
\caption{Comparison with different feature pyramid networks on MS-COCO val2017. The symbol `*' represents the result of our re-implementation, with experimental details consistent with the proposed method. The backbone used in all the listed methods is ResNet-50. The best and second-best results of models with similar input size are marked with violet and purple respectively.}
\centering
\resizebox{0.85\linewidth}{!}
{
\begin{tabular}{lccccccccc} 
\toprule
\textbf{Method}                                 & \textbf{Image size}  & \textbf{AP}                    & \textbf{AP$_{\bm{50}}$}       & \textbf{AP$_{\bm{75}}$}       & \textbf{AP$_{\bm{S}}$}        & \textbf{AP$_{\bm{M}}$}        & \textbf{AP$_{\bm{L}}$}        & \textbf{Params}   & \textbf{GFLOPs}   \\ 
\midrule
Faster R-CNN+FPN*~\cite{lin2017feature}         & $640 \times 640$     & 37.4                           & 57.3                          & 40.3                          & 18.4                          & 41.7                          & 52.7                          & 41.5 M            & 91.4              \\
Faster R-CNN+PAFPN*~\cite{liu2018path}          & $640 \times 640$     & \color{violet}\textbf{38.1}    & \color{violet}\textbf{58.1}   & \color{violet}\textbf{41.3}   & \color{violet}\textbf{19.1}   & 42.5                          & 54.0                          & 45.1 M            & 101.3             \\
Faster R-CNN+NASFPN*~\cite{ghiasi2019fpn}       & $640 \times 640$     & 37.7                           & 54.5                          & 41.1                          & 15.5                          & \color{violet}\textbf{44.5}   & \color{violet}\textbf{56.9}   & 68.2 M            & 103.0             \\
Faster R-CNN+CARAFE~\cite{wang2019carafe}       & $800 \times 1333$    & 38.6                           & 59.9                          & 42.2                          & 23.3                          & 42.2                          & 49.7                          & 47.1 M            & 219.8             \\
Faster R-CNN+AugFPN~\cite{guo2020augfpn}        & $800 \times 1333$    & 38.7                           & \color{purple}\textbf{61.2}   & 41.9                          & \color{purple}\textbf{24.1}   & 42.5                          & 49.5                          & --                & --                \\
Mask R-CNN+FPT~\cite{zhang2020feature}          & $800 \times 1000$    & 38.0                           & 57.1                          & 38.9                          & 20.5                          & 38.1                          & \color{purple}\textbf{55.7}   & 88.2 M            & 346.2             \\
Faster R-CNN+DRFPN~\cite{ma2020dual}            & $800 \times 1333$    & 39.1                           & 60.3                          & 42.5                          & 22.9                          & \color{purple}\textbf{43.1}   & 50.7                          & 48.4 M            & 263.0             \\
Faster R-CNN+GraphFPN~\cite{zhao2021graphfpn}   & $800 \times 1000$    & 39.1                           & 58.3                          & 39.4                          & 22.4                          & 38.9                          & \color{purple}\textbf{56.7}   & 100.0 M      & 380.0             \\
Faster R-CNN+LFPN~\cite{xie2022latent}          & $800 \times 1333$    & 38.7                           & \color{purple}\textbf{60.4}   & 41.9                          & 23.5                          & 42.5                          & 49.0                          & --                & --                \\
Faster R-CNN+ImFPN~\cite{zhu2022improved}       & $800 \times 1333$    & \color{purple}\textbf{39.2}    & 59.9                          & \color{purple}\textbf{42.6}   & 22.2                          & 42.8                          & 52.1                          & --                & --                \\
\midrule
Faster R-CNN+AFPN                               & $640 \times 640$     & \color{violet}\textbf{39.0}    & \color{violet}\textbf{57.6}   & \color{violet}\textbf{42.1}   & \color{violet}\textbf{19.4}   & \color{violet}\textbf{43.0}   & \color{violet}\textbf{55.0}   & 50.2 M            & 90.0              \\
Faster R-CNN+AFPN (CARAFE)                      & $640 \times 640$     & 39.2                           & 57.8                          & 42.3                          & 19.9                          & 43.2                          & 55.5                          & 52.2 M            & 92.5              \\
Faster R-CNN+AFPN                               & $800 \times 1000$    & \color{purple}\textbf{41.0}    & 60.3                          & \color{purple}\textbf{44.2}   & \color{purple}\textbf{23.7}   & \color{purple}\textbf{44.8}   & 53.0                          & 50.2 M            & 165.6             \\
Faster R-CNN+AFPN (CARAFE)                      & $800 \times 1000$    & 41.9                           & 61.3                          & 45.4                          & 24.7                          & 45.6                          & 54.2                          & 52.2 M            & 170.8             \\
\bottomrule
\end{tabular}
}
\label{tab1}
\end{table*}

\begin{table}[htbp]
\caption{Comparison with different feature pyramid networks on MS-COCO test-dev. The backbone used in all listed methods is ResNet-101, and the input image size is $800 \times 1000/1333$ pixels.}
\centering
\resizebox{\linewidth}{!}
{
\begin{tabular}{lccccccccc} 
\toprule
\textbf{Method}                                 & \textbf{AP}   & \textbf{AP$_{\bm{50}}$}   & \textbf{AP$_{\bm{75}}$}   & \textbf{AP$_{\bm{S}}$}    & \textbf{AP$_{\bm{M}}$}    & \textbf{AP$_{\bm{L}}$} \\ 
\midrule
Faster R-CNN+FPN~\cite{lin2017feature}          & 39.7          & 61.4                      & 43.3                      & 22.3                      & 42.9                      & 50.4                   \\
Faster R-CNN+AugFPN~\cite{guo2020augfpn}        & 41.5          & \textbf{63.9}             & 45.1                      & 23.8                      & 44.7                      & 52.8                   \\
Mask R-CNN+FPT~\cite{zhang2020feature}          & 41.6          & 60.9                      & 44.0                      & 23.4                      & 41.5                      & 53.1                   \\
Faster R-CNN+DRFPN~\cite{ma2020dual}            & 41.8          & 63.0                      & 45.7                      & 23.1                      & 44.7                      & 53.1                   \\
Faster R-CNN+GraphFPN~\cite{zhao2021graphfpn}   & 42.1          & 61.3                      & \textbf{46.1}             & 23.6                      & 41.1                      & 53.3                   \\
RetinaNet+LFPN~\cite{xie2022latent}             & 40.0          & 60.3                      & 42.8                      & 22.6                      & 43.2                      & 50.5                   \\
Faster R-CNN+ImFPN~\cite{zhu2022improved}       & 41.4          & 61.9                      & 45.2                      & 22.8                      & 44.5                      & 53.1                   \\
\midrule
Faster R-CNN+AFPN                               & \textbf{42.3} & 61.8                      & 46.0                      & \textbf{24.6}             & \textbf{45.3}             & \textbf{53.8}          \\
\bottomrule
\end{tabular}
}
\label{tab2}
\end{table}

\subsection{Asymptotic Architecture}

The architecture of the proposed AFPN is illustrated in Fig.~\ref{fig:1}.
During the bottom-up feature extraction process of the backbone network, AFPN asymptotically integrates low-level, high-level, and top-level features. Specifically, AFPN initially fuses low-level features, followed by deep features, and finally integrates the topmost features, i.e., the most abstract ones. The semantic gap between non-adjacent hierarchical features is larger than that between adjacent hierarchical features, especially for the bottom and top features. This leads to the poor fusion effect of non-adjacent hierarchical features directly. Therefore, it is unreasonable to directly use $C_{2}$, $C_{3}$, $C_{4}$ and $C_{5}$ for feature fusion. Since the architecture of AFPN is asymptotic, this will make the semantic information of different levels of features closer in the process of asymptotic fusion, thus alleviating the above problems. For example, the feature fusion between $C_{2}$ and $C_{3}$ reduces their semantic gap. Since $C_{3}$ and $C_{4}$ are adjacent hierarchical features, the semantic gap between $C_{2}$ and $C_{4}$ is reduced.


To align the dimensions and prepare for feature fusion, we utilize $1 \times 1$ convolution and bilinear interpolation methods for upsampling the features. On the other hand, we perform downsampling using different convolutional kernels and strides depending on the required downsample rate. For instance, we apply a $2 \times 2$ convolution with a stride of 2 to achieve 2 times downsampling, a $4 \times 4$ convolution with a stride of 4 for 4 times downsampling, and an $8 \times 8$ convolution with a stride of 8 for 8 times downsampling. Following feature fusion, we continue learning features using four residual units, which are similar to ResNet~\cite{he2016deep}. Each residual unit comprises two $3 \times 3$ convolutions. Due to the use of only three levels of features in YOLO, there is no 8 times upsampling and 8 times downsampling.

\subsection{Adaptive spatial fusion}

We leverage ASFF~\cite{liu2019learning} to assign varying spatial weights to the different levels of features during the multi-level feature fusion process, enhancing the significance of pivotal levels and mitigate the impact of contradictory information from different objects. As depicted in Fig.~\ref{fig:2}, we fuse the features of the three levels. Let $x_{ij}^{n \rightarrow l}$ denote the feature vector at position $(i, j)$ from level $n$ to level $l$. The resultant feature vector, denoted as $y_{ij}^{l}$, is obtained through the adaptive spatial fusion of multi-level features and is defined by the linear combination of feature vectors $x_{ij}^{1 \rightarrow l}$, $x_{ij}^{2 \rightarrow l}$, and $x_{ij}^{3 \rightarrow l}$ as follows: 
\begin{equation}
y_{ij}^{l} = \alpha_{ij}^{l} \cdot x_{ij}^{1 \rightarrow l} + \beta_{ij}^{l} \cdot x_{ij}^{2 \rightarrow l} + \gamma_{ij}^{l} \cdot x_{ij}^{3 \rightarrow l},
\label{eq1}
\end{equation}
where $\alpha_{ij}^{l}$, $\beta_{ij}^{l}$, and $\gamma_{ij}^{l}$ represent the spatial weights of the features of the three levels at level $l$, subject to the constraint that $\alpha_{ij}^{l}$ + $\beta_{ij}^{l}$ + $\gamma_{ij}^{l} = 1$. Given the discrepancy in the number of fused features at each stage of the AFPN, we implement a stage-specific number of adaptive spatial fusion modules.

\begin{table*}[htbp]
\caption{Comparison with other two-stage target detectors. The 1x in the Schedule represents training 12 epochs, and the 2x represents training 24 epochs. The AP on the left is evaluated on val2017, and the AP on the right is evaluated on test-dev. The backbone used in all the listed methods is ResNet-50.}
\centering
\resizebox{\linewidth}{!}
{
\begin{tabular}{l|c|cccccc|cccccc} 
\toprule
\textbf{Method}                                 & \textbf{Schedule}  & \textbf{AP}   & \textbf{AP$_{\bm{50}}$}   & \textbf{AP$_{\bm{75}}$}   & \textbf{AP$_{\bm{S}}$}    & \textbf{AP$_{\bm{M}}$}    & \textbf{AP$_{\bm{L}}$}    & \textbf{AP}   & \textbf{AP$_{\bm{50}}$}   & \textbf{AP$_{\bm{75}}$}   & \textbf{AP$_{\bm{S}}$}    & \textbf{AP$_{\bm{M}}$}    & \textbf{AP$_{\bm{L}}$}   \\ 
\midrule
Faster R-CNN+FPN~\cite{lin2017feature}          & 1x   & 37.4   & 58.1  & 40.4  & 21.2  & 41.0  & 48.1  & 37.7  & 58.7  & 40.8  & 21.7  & 40.6  & 46.7       \\
Faster R-CNN+AFPN                               & 1x   & 38.6   & 57.7  & 41.7  & 21.5  & 42.1  & 51.3  & 38.8  & 58.5  & 41.9  & 21.4  & 41.5  & 49.0       \\
\midrule
Faster R-CNN+FPN~\cite{lin2017feature}          & 2x    & 38.4  & 59.0  & 42.0  & 21.5  & 42.1  & 50.3  & 38.7  & 59.6  & 42.1  & 22.1  & 41.4  & 48.6       \\
Faster R-CNN+AFPN                               & 2x    & 38.9  & 57.4  & 42.0  & 21.3  & 42.0  & 51.3  & 39.3  & 58.4  & 42.5  & 21.4  & 41.6  & 50.1       \\
\midrule
Dynamic R-CNN+FPN~\cite{zhang2020dynamic}       & 1x    & 38.9  & 57.6  & 42.7  & 22.1  & 41.9  & 51.7  & 39.2  & 58.3  & 42.9  & 22.1  & 41.9  & 49.6       \\
Dynamic R-CNN+AFPN                              & 1x    & 39.5  & 56.9  & 42.8  & 20.7  & 42.9  & 53.4  & 39.8  & 57.5  & 43.3  & 21.5  & 42.2  & 51.3       \\
\bottomrule
\end{tabular}
}
\label{tab3}
\end{table*}

\begin{table}[htbp]
\caption{Contribution of our AFPN to YOLOv5.}
\centering
\resizebox{\linewidth}{!}
{
\begin{tabular}{lccccccc} 
\toprule
\textbf{Method} & \textbf{Neck}             & \textbf{AP}   & \textbf{AP$_{\bm{S}}$}    & \textbf{AP$_{\bm{M}}$}    & \textbf{AP$_{\bm{L}}$}    & \textbf{Params}    \\ 
\midrule
YOLOv5-n        & YOLOv5PAFPN~\cite{yolov5} & 28.0          & 14.0                      & 31.8                      & 36.6                      & 1.87 M             \\                  
YOLOv5-n        & AFPN                      & 29.3          & 14.2                      & 32.2                      & 40.0                      & 1.67 M             \\                   
\midrule
YOLOv5-s        & YOLOv5PAFPN~\cite{yolov5} & 37.7          & 21.7                      & 42.5                      & 48.8                      & 7.24 M             \\                   
YOLOv5-s        & AFPN                      & 38.6          & 22.1                      & 42.7                      & 51.4                      & 6.42 M             \\                  
\bottomrule
\end{tabular}
}
\label{tab4}
\end{table}

\begin{table}[htbp]
\caption{Ablation studies on the fusion operation.}
\centering
\resizebox{\linewidth}{!}
{
\begin{tabular}{lccccccccc} 
\toprule
\textbf{Fusion Operation}       & \textbf{AP}       & \textbf{AP$_{\bm{50}}$}   & \textbf{AP$_{\bm{75}}$}   & \textbf{AP$_{\bm{S}}$}    & \textbf{AP$_{\bm{M}}$}    & \textbf{AP$_{\bm{L}}$}    \\ 
\midrule
concat                          & 38.8              & 57.3                      & 42.0                      & \textbf{19.6}             & 42.7                      & 54.6                      \\
sum                             & 38.5              & 57.2                      & 41.6                      & 18.8                      & 42.6                      & 54.1                      \\
adaptive spatial fusion         & \textbf{39.0}     & \textbf{57.6}             & \textbf{42.1}             & 19.4                      & \textbf{43.0}             & \textbf{55.0}             \\
\bottomrule
\end{tabular}
}
\label{tab5}
\end{table}

\section{Experiments} 

\subsection{Experiment Setup}

\paragraph*{\textbf{Datasets}}
We evaluate the proposed method on the MS COCO 2017 dataset~\cite{lin2014microsoft}, 118k training images (train2017), 5k validation images (val2017), and 20k testing images (test-dev). Due to the unavailability of the test-dev labels, we uploaded the model-generated bounding box to a designated evaluation website to obtain performance metrics. Specifically, we select average precision (AP), AP$_{50}$, AP$_{75}$, AP$_{S}$, AP$_{M}$, and AP$_{L}$ as the evaluation metrics. 

\paragraph*{\textbf{Implementation Details}}
We utilize the MMDetection~\cite{mmdetection} as the underlying framework and conduct the experiments on 2 NVidia RTX3090 GPUs. During training, we adopt SGD~\cite{loshchilov2016sgdr} as our optimizer and configure the learning rate, weight decay, and momentum to be 0.01, 0.0001, and 0.9, respectively. Each mini-batch contains 8 images distributed on 2 GPUs. For fair comparison, we used images with different resolutions as input in different experiments. We will describe the specific situation in each comparative experiment part. The remaining hyperparameters follow the default configuration of MMDetection.

\subsection{Comparison with Different Feature Pyramid Networks}

In this section, the methods denoted with `*' and our proposed methods were trained for 36 epochs. The learning rate was reduced by a factor of 10 at the 27-th and 33-th epochs, respectively. Random flipping and random cropping are used in the data augmentation process. We compare the performance of our method with recent feature pyramid networks. Given that the model’s performance is heavily dependent on the input image size, we conduct a comparative analysis of our proposed method and recent feature pyramid networks using input images of similar resolution. 

As shown in Table~\ref{tab1}, our method achieves strong performance with an AP of 39.0\% when the input image size is $640 \times 640$, even surpassing that of some larger resolution models. Compared to FPN~\cite{lin2017feature} and PAFPN~\cite{liu2018path}, our AFPN demonstrated improvement in AP performance on val2017 by 1.6\% and 0.9\%, respectively, while surpassing them on most other metrics. It is worth noting that, as NASFPN~\cite{ghiasi2019fpn} is searched on the RetinaNet~\cite{lin2017focal} framework, its performance on the Faster R-CNN~\cite{ren2015faster} framework is not particularly impressive. In contrast, our AFPN outperforms NASFPN by 1.3\% in AP. Our AFPN achieves an AP of 41.0\% when the input image size is $800 \times 1000$, surpassing the performance of other methods. In constructing the AFPN architecture, we did not take into account the quality of upsampling. To address this deficiency, we have replaced the bilinear interpolation operator with the CAFAFE~\cite{wang2019carafe} operator, which boasts superior upsampling quality. In conducting further experiments, we found that this substitution led to a notable enhancement in the performance of our model. Moreover, we replace the backbone with ResNet-101~\cite{he2016deep} for training and testing on MS COCO test-dev~\cite{lin2014microsoft}. Table~\ref{tab2} demonstrates a 2.6\% increase in AP of our AFPN compared to the baseline (FPN). Our method also achieves competitive results when compared with similar techniques while maintaining a leading position in AP, AP$_{S}$, AP$_{M}$, and AP$_{L}$.

\subsection{Results on Different Detectors}

To demonstrate the versatility of our method, we incorporated our AFPN into both two-stage and one-stage detectors. Experimental results indicate that our method significantly enhances the performance of both detector frameworks.

\paragraph*{\textbf{Two-stage Detectors}}
The experimental results on the two-stage detectors are shown in Table~\ref{tab3}. The input image size of all methods in the table is $800 \times 1333$. Only random flips are used in the data augmentation process. We evaluated Faster R-CNN~\cite{lin2017feature} and Dynamic R-CNN~\cite{zhang2020dynamic} in our study. Our experimental results demonstrated that under the same training time, replacing the FPN of the detector with AFPN can significantly enhance the detection performance, particularly for the detection of large objects. This is because the architecture of FPN does not allow high-level features to obtain detailed information of low-level features. Our AFPN does not improve the detector's ability to detect small targets, which is supported by the AP$_{S}$ results. Moreover, we also found that AFPN is inferior to FPN in AP$_{50}$ but superior to FPN in AP$_{75}$. Therefore, compared to FPN, our AFPN is more suitable for high-precision positioning scenarios.

\paragraph*{\textbf{One-stage Detectors}}
The experimental results on the YOLOv5~\cite{yolov5} are shown in Table~\ref{tab4}. The detector was trained for 300 epochs using input images of size $640 \times 640$. Our experimental results demonstrate a significant improvement in detection performance with our AFPN compared to the original neck (YOLOv5PAFPN) of YOLOv5, particularly for detecting large objects. Specifically, for YOLOv5-n, our AFPN improves the average precision of large objects (AP$_{L}$) by 3.4\%, and for YOLOv5-s, it improves AP$_{L}$ by 2.6\%. Furthermore, our AFPN maintains a leading position in AP, AP$_{S}$, AP$_{M}$, and AP$_{L}$.

\subsection{Learnable Parameters and Computational Cost}

Both the depth and width of the network can affect its representation ability. As the depth of the AFPN has already enhanced the model's representation ability, we adopted a strategy of reducing the network width to optimize the model. Specifically, in the two-stage detector, we reduced the dimension of the features entering the feature pyramid network to the original $1/8$. And in the one-stage detector, we reduced the dimension to the original $1/4$. Table~\ref{tab1} provides the number of learnable parameters and total computational cost of various feature pyramid networks, including our proposed AFPN. Based on the results presented in the table, our AFPN architecture has 50.2 million learnable parameters, and GFLOPs achieve 90.0 at the resolution of $640 \times 640$. Compared to FPN~\cite{lin2017feature}, the number of parameters in our AFPN increased by 21.0\%. However, we achieve the lowest GFLOPs among all the methods in the table. The main reason for this phenomenon is that we reduce the feature dimension. The experimental results in Table~\ref{tab4} show that AFPN achieves improved performance on YOLOv5 while utilizing fewer parameters.

\subsection{Ablation Studies}

To investigate the efficacy of adaptive spatial fusion operation in our AFPN, we replaced it with two other fusion operations, namely element-wise sum and element-wise concatenation, for ablation studies. Our experimentation utilized the Faster R-CNN framework with ResNet-50 as the backbone. As indicated in Table~\ref{tab5} of our ablation study, we observed that the element-wise concatenation operation could attain performance levels comparable to those achieved with adaptive spatial fusion operation. However, the AP, AP$_{50}$, AP$_{75}$, AP$_{M}$, and AP$_{L}$ metrics were slightly lower. Given that adaptive spatial fusion performs a weighted operation on the element-wise sum to suppress contradictions between features, it is reasonable to assume that it would perform better than the element-wise sum. The experimental results also prove this.

\section{Conclusion}

In this paper, we propose the Asymptotic Feature Pyramid Network (AFPN) to solve the problem of information loss and degradation caused by indirect interaction between non-adjacent levels. Our AFPN uses an asymptotic way for feature fusion and adaptive spatial fusion operation to extract more useful information during the fusion process. Extensive experimental results demonstrate the superior performance of AFPN when compared to baseline methods across various detection frameworks. In the future, we will explore a lighter AFPN and its applicability in other visual tasks.

\section{Acknowledgement}

This work was supported in part by the National Natural Science Foundation of China (No. 62036009, No. 62106226), the National Key Research and Development Program of China (No. 2020YFB1707700), and Zhejiang Provincial Natural Science Foundation of China (No. LQ22F020013, No.LDT23F0202, No. LDT23F02021F02).

\bibliographystyle{IEEEtran}
\bibliography{sample}

\end{document}